# Improving Fabrication Fidelity of Integrated Nanophotonic Devices Using Deep Learning

## Authors

Dusan Gostimirovic[a*], Yuri Grinberg[b], Dan-Xia Xu[c], Odile Liboiron-Ladouceur[a]

[a]Department of Electrical and Computer Engineering, McGill University, 845 Sherbrooke Street West, Montreal, H3A 0G4, Canada

[b]Digital Technologies Research Centre, National Research Council Canada, 1200 Montreal Road, Ottawa, K1A 0R6, Canada

[c]Advanced Electronics and Photonics Research Centre, National Research Council Canada, 1200 Montreal Road, Ottawa, K1A 0R6, Canada

Email: dusan.gostimirovic@mcgill.ca

## Abstract

Next-generation integrated nanophotonic device designs leverage advanced optimization techniques such as inverse design and topology optimization which achieve high performance and extreme miniaturization by optimizing a massively complex design space enabled by small feature sizes. However, unless the optimization is heavily constrained, the generated small features are not reliably fabricated, leading to optical performance degradation. Even for simpler, conventional designs, fabrication-induced performance degradation still occurs. The degree of deviation from the original design not only depends on the size and shape of its features, but also on the distribution of features and the surrounding environment, presenting complex, proximity-dependent behavior. Without proprietary fabrication process specifications, design corrections can only be made after calibrating fabrication runs take place. In this work, we introduce a general deep machine learning model that automatically corrects photonic device design layouts prior to first fabrication. Only a small set of scanning electron microscopy images of engineered training features are required to create the deep learning model. With correction, the outcome of the fabricated layout is closer to what is intended, and thus so too is the performance of the design. Without modifying the nanofabrication process, adding significant computation in design, or requiring proprietary process specifications, we believe our model opens the door to new levels of reliability and performance in next-generation photonic circuits.

# Introduction

High speed, low power consumption, and adherence to existing nanofabrication materials and processes make silicon photonics one of the ideal candidates for the push of "beyond Moore" in next-generation heterogenous computing and communications[1–4]. The major drawback of silicon photonics—beyond its relatively poor lasing and signal modulation capabilities—is its sensitivity to fabrication process induced structural deviations[5–7]. As device designs become more compact and complex to further push performance, this fabrication sensitivity becomes even more severe. New and revolutionary design methodologies such as topology optimization achieve designs with promising levels of performance and miniaturization[8–11]; however, the fine and complex structural features of these designs are not reliably fabricated without including additional design constraints[12–15]. This trade-off between performance, miniaturization, and fabrication robustness is currently balanced by constraining the design optimization to only generate features that satisfy the design rule constraints specified by the nanofabrication facility (e.g., minimum feature sizing, spacing, and curvature), or by reducing the amount of light interaction with material interfaces altogether. These types of constraints, although generally effective in ensuring an acceptable layout submission for fabrication, do not capture the full relationship between the nanofabrication capabilities and the complex shapes and arrangements of the design features. Even for simple, conventional photonic designs, adhering to design rule constraints does not guarantee perfect design fidelity, nor does it push the limits of the existing technology. Thus, it is desirable to efficiently capture the relationship between the device layout and its fabricated outcome so that a designer can apply the constraints more precisely and pre-compensate for the anticipated deviations.

Our previous work introduced the use of deep convolutional neural networks (CNNs) trained to learn the relationship between graphic design system (GDS) layouts and scanning electron microscope (SEM) images so they can be used to predict fabrication deviations in planar silicon photonic devices[16], as shown by the "Previous Work" box in Figure 1. Major deviations of over-etched convex bends, under-etched concave bends, loss of small islands, and filling of narrow holes/channels are accurately predicted; and the fabrication variance (represented by the uncertainty of the neural network model predictions) is characterized in this virtual fabrication environment. The anticipated optical performance can then be simulated using the predicted structure. This method can be used to characterize a range of design options without costly and lengthy prototyping fabrication runs. Nevertheless, because of the complex relationship between a photonic device design layout and its predicted fabrication deviations, it is desirable to make automatic corrections to the layout so that the fabricated outcome is more like the desired structure as designed.

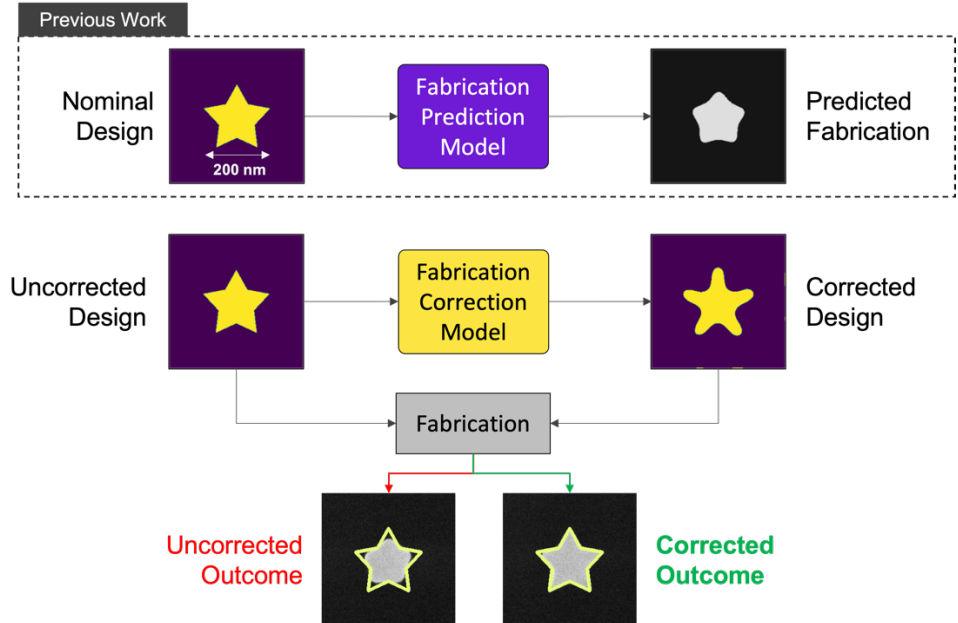

*Figure 1. Overview of the proposed fabrication variation prediction and correction models based on deep machine learning. A CNN is trained to work as a predictor model (our previous work[16]). A second CNN is trained to work as a corrector model that modifies a silicon photonic design layout so that its fabricated outcome is closer to the original design.*

In this work, we present a new model for the *automated correction* of fabrication deviations in planar silicon photonic devices, also based on deep convolutional neural networks, as shown by the flowchart in Figure 1. We propose to use a tandem network architecture for the model training, which has been shown to improve the accuracy of similar design problems[17]. The presented corrector model adds silicon where it expects to lose silicon, and vice versa, so that the fabricated outcome, and hence the optical performance, is closer to that of the intended, original design. Although the concept of reshaping design layouts to correct for fabrication deviations can be achieved in existing inverse lithography techniques such as shape proximity effect correction[18–20], such work so far has focused on Manhattan-like shapes used in microelectronics to correct for electronic failure modes, and it is not evident how well they work for complex, topologically optimized photonic device designs. Furthermore, conventional inverse lithography techniques may require proprietary specifications about the nanofabrication process and are therefore generally not available to designers that outsource their fabrication (i.e., through multi-project wafer runs). Our approach further differentiates itself by capturing the effects of the entire fabrication process through data, rather than mathematically modelling each process step. Training our model only requires a modest set of readily available SEM images to train, does not modify the existing fabrication process, and does not add significant computation to the design process. The model enables "free" improvement of any planar silicon photonic device design, but it can also be used to relax the fabrication constraints in future designs knowing that the model can restore the predicted deviations. For usage and testing of our correction model, and to stimulate further research on this topic, our code repository is made available on GitHub[21].

The remainder of this paper is organized as follows. The Methods section begins by summarizing the preparation, acquisition, and processing of the training data. The same data processing was used in our previous work on CNN-based fabrication deviation prediction and is directly carried over to this work on the correction of such predicted deviations. The Methods section also outlines the construction of our deep tandem CNN corrector model training structure, how it is trained, how it is inferenced to produce corrections, and how to interpret the corrections. The Results and Discussion section begins by presenting the correction of simple structures to demonstrate the capability of our model to improve the fidelity of fine planar silicon photonic features. It then concludes with the example of the correction of a complex, fine-featured, inverse-designed wavelength-division (de)multiplexer (WDM) device, which shows a significant level of optical performance improvement our model can restore.

## Methods

Given a well-performing fabrication predictor model, as was verified in our previous work[16], we train a corrector model on the same fabrication process. The predictor and corrector models are complementary and are thus both used in this work. As shown in Figure 1, and assuming a stable and unchanged fabrication process, the task of the corrector model is to modify the layout of a design such that the predicted deviations will revert its form back to what was first intended. Like the predictor, the corrector model learns the complex relationship between design and fabrication, but now in the reverse order. That is, the input to the corrector model is the desired design to be realized in fabrication, and its output is the corrected layout to be submitted for fabrication. For clarity in terms, *nominal* is the desired form; *prediction* refers to the form that is predicted to result from fabricating a given nominal layout; *correction* is the layout to be fabricated to obtain the nominal form after fabrication; and *outcome* (predicted or fabricated) is the form of the predicted or fabricated correction. The terms, *layout*, *structure*, *device,* and *form* are used interchangeably throughout different contexts.

### Preparation of Training Data

The data preparation process for training the proposed corrector model is the same as what was used in our previous work for the predictor model[16]. A set of 30 geometrical patterns (3.0×2.25 μm$^2$ in size), as shown in Figure 2, are fabricated with the NanoSOI electron-beam lithography process from Applied Nanotools Inc.[22], which are then imaged using SEM. These patterns were generated to contain a range of features that are representative of those found in next-generation (inverse) photonic device designs but are still suitable to train a model that can be used on conventional photonic device designs as well. To capture the full variability of the nanofabrication process, data can be obtained from multiple runs and trained together. A data preprocessing stage matches each SEM to its corresponding GDS by resizing, aligning, and binarizing. The details of the preprocessing stage can be found in the Supplementary Information of Ref. [16]. The 2048×1536 pixel$^2$ SEM images are sliced into overlapping 128×128 pixel$^2$ slices to (i) reduce the computational load in training (i.e., to limit the size of the neural network), (ii) to artificially create more training data, and (iii) to create a more flexible model that corrects devices of any shape and size through many smaller corrections. Based on the SEM imaging size, the resolution of the model in this work is approximately 1.5 pixel/nm. For a finer resolution, the SEM imaging can be taken using a higher magnification, but the imaging area will be reduced, and more images will be required to gather the same amount of information.

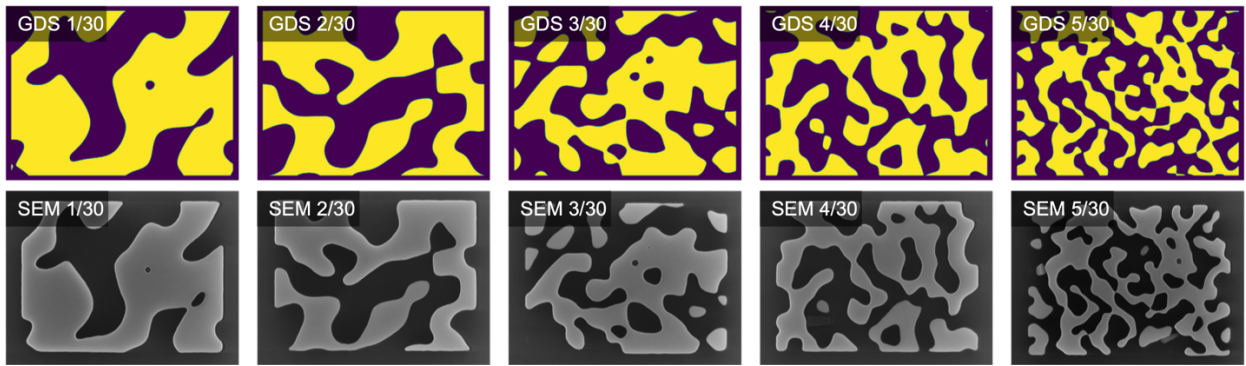

*Figure 2. Five example training patterns (GDS) and their corresponding SEM images. The patterns are randomly generated to create many different feature types and feature distributions to train a well-generalized convolutional neural network model.*

### Corrector Model Training

While our predictor model learned the translation from designs (GDS) to fabricated structures (SEM), in this work, the corrector model learns the *inverse* translation from fabricated structures to original designs. This distinction is visualized in Figure 3a and Figure 3b. As previously visualized in Figure 1, the usage (inference) of the trained corrector model has a desired outcome inputted (i.e., the nominal design), and a corrected layout is generated at its output. The fabricated outcome of this corrected layout is intended to closely resemble the nominal design. To achieve this goal, we train an inverse model by inputting the SEM slices, and the GDS slices are used as ground-truth labels to check the error of the generated output (a correction). No other changes are required to achieve this functionality.

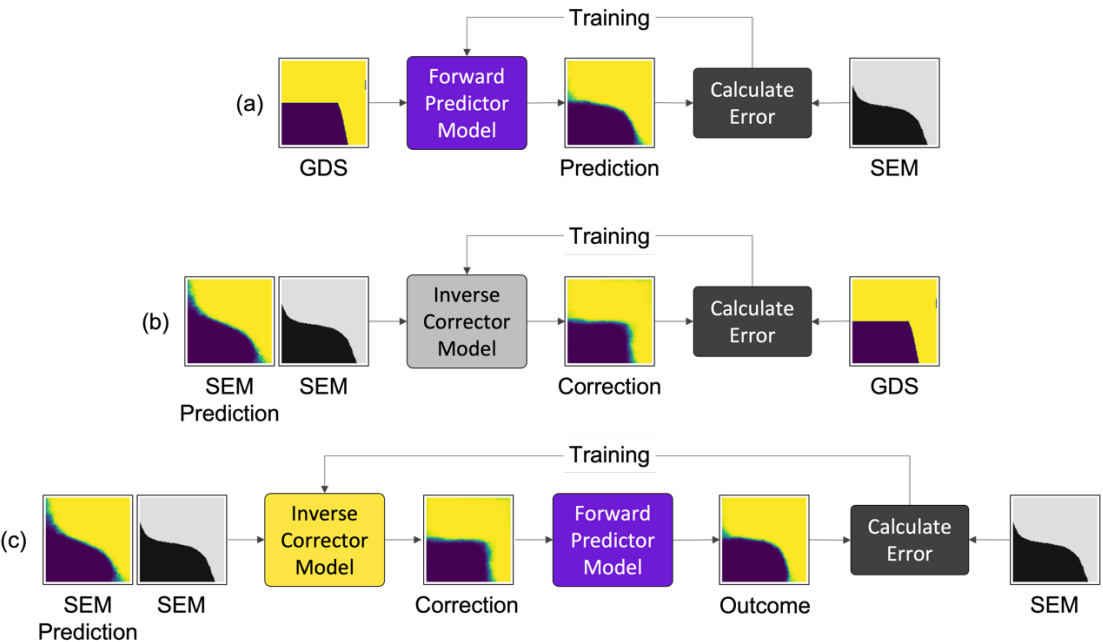

*Figure 3. Process diagrams for the training of the (a) forward predictor model, the (b) inverse corrector model trained independently, and the (c) inverse corrector model trained in a tandem architecture. Prediction of the input is shown in (b) and (c) to show the expected feature smoothing without correction.*

The predictor model, also called here a forward model, is established through a standard supervised learning process, and we have demonstrated its high accuracy. However, building an inverse model by swapping inputs with outputs—as described above and illustrated in Fig. 3b—loses accuracy as there may be many valid solutions (corrections) to one input (nominal design). Our neural network is trained to minimize classification errors across 10,000 test examples: if a particular example can be classified correctly in multiple ways, the neural network will be fed conflicting information and will struggle to converge. This is known as a one-to-many mapping problem, which is especially acute in inverse problems. To overcome this difficulty, we adopt a tandem neural network training structure that is known to alleviate this problem by attaching a pretrained forward model to the output of the to-be-trained inverse model[17], as shown in Figure 3c. The to-be-trained inverse model accepts a batch $N$ of 128×128 pixel$^2$, single-channel examples (an $N$×128×128×1 tensor) at its input and produces a same-size correction at its output. The attached pre-trained forward model at the output acts as a "decision circuit" to bias the inverse model towards one of the many solutions for each type of example. While potentially many good solutions are discarded in the process, only one good solution is needed for the model to be effective in generating useful corrections. Figure 4 shows the structure of the inverse model that can be trained independently, or as part of the tandem neural network architecture.

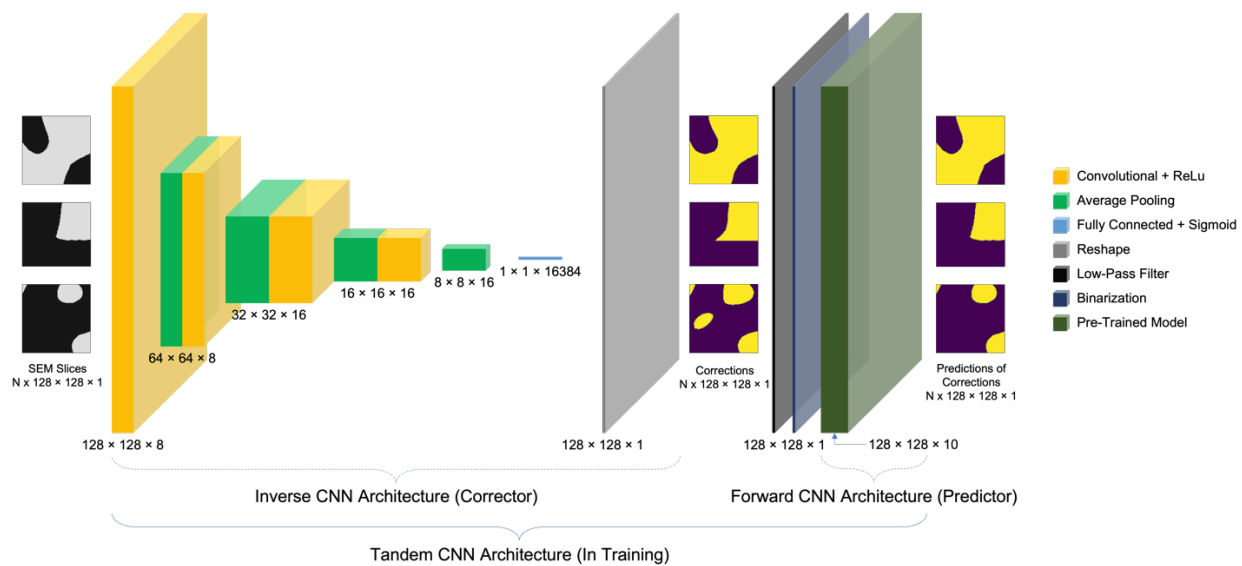

*Figure 4. The structure of the proposed tandem CNN architecture used for training. An inverse CNN model is trained to produce design layout corrections, which are inputted into a pre-trained CNN predictor model to produce predictions of corrections. Those are then compared with ground-truth SEM labels to calculate the error of the model and update the CNN weights. After training, only the inverse model is used to produce corrections through inference.*

The model in Figure 4 is constructed and trained using the open-source machine learning library, TensorFlow[23]. For the inverse model, four convolutional layers are connected in series (channel sizing of 8, 8, 16, then 16), each with average pooling (using a 2×2 pixel$^2$ kernel size) for dimensionality reduction, and ReLu activation for nonlinearity. At the output of the final convolutional layer is a single fully connected layer with a sigmoid activation and a reshaping layer that maps the convolutions back into a 128×128 pixel$^2$ output (correction). When trained independently, the output is compared with its corresponding GDS slice in training and the weights are updated using backpropagation.

The structure of the inverse model itself remains unchanged for a fair comparison of both training approaches in this work. The training process proceeds by updating the weights of the inverse model until the inputs (ground-truth SEMs) match the outputs (predictions of corrections) as closely as possible, as shown in Figure 3c. For further accuracy improvements, the pre-trained forward model of this tandem network is an ensemble model, which is a collection of 10 identically structured forward models that are trained with different random weight initializations and dataset ordering[24,25]. This improves the accuracy of the predictor model, and thus enhances its role in the tandem training architecture. The networks are trained with the adaptive moment estimation optimization method (Adam) and the binary cross-entropy (BCE) loss function. With BCE, for each pixel of an inputted SEM image, the corrector model classifies the probability of the corresponding pixel of the correction being silicon. Our dataset is split into training and testing subsets with an 80:20 randomized distribution. The model stops training when the BCE for a set of unseen, testing data is minimized—indicating high certainty in the model's correction. The corrector model trained in a tandem architecture achieves a BCE of 0.045, indicating that 4.5% of the pixels in the testing data are uncertain. The BCE for the inverse model trained independently is twice as large as the above, at BCE = 0.091.

Making Corrections

For inference (usage) of the corrector model trained with a tandem architecture, the pretrained forward model that was used in training is removed. As the corrector model is trained on small, 128×128 pixel$^2$ slices, it can only make corrections over a small area. Therefore, a full device design is corrected by making corrections over many small areas and stitching them together. This approach is common to other application of deep convolutional neural networks[26]. To increase accuracy and correction smoothness, an overlapping stitch step size is used, where multiple corrections can be made from multiple perspectives and averaged together. Furthermore, an ensemble of 10 identically structured corrector models, with different random initializations of the weights and training dataset ordering, are used in each correction to further reduce training biases and increase overall correction accuracy.

Figure 5 shows an illustrative example of a structure to be corrected: a simple cross with 200×50 nm$^2$ crossings in a 256×256 pixel$^2$ image. A four-pixel scanning step size is used for an ultrahigh-quality result, at the expense of computation time (though still only 13 s to complete the full correction). The prediction of the cross in Figure 5b has its corners rounded—with over-etching of convex corners (silicon inside the corner) and under-etching of concave corners (silicon outside the corner). The correction of the cross in Figure 5c adds silicon where it expects to lose silicon, and vice versa, creating an exaggerated cross shape significantly different than the nominal. Note that the raw outputs of the predictor and corrector models are not binary; there are regions around the edges of the structure that are neither silicon nor silica. These regions represent the uncertainty of the model, which arises from imperfections in the training process and minor variations in the fabrication process stemming from spatial changes across the wafer and time-varying conditions in patterning. Therefore, a well-trained predictor model predicts the major deterministic variations in the design (e.g., corner rounding) and the statistical uncertainty of where an edge may lie from device to device. Likewise, a well-trained corrector model will correct the major variations, but the edges still may vary from device to device, within the bounds of the uncertainty region. For this demonstration, the half-way point of the uncertainty region can be taken as the most likely location of the edge, and the structure can be binarized there. When the correction is predicted using the forward model, the outcome in Figure 5d is much closer to the desired nominal. There are 1,401 error pixels between nominal and prediction versus 586 error pixels between nominal and prediction of correction—a reduction factor of 2.4. These differences are visualized in Figure 5h–j. The corners of Figure 5j still have a small degree of rounding: this is in part because the model is trained on a dataset that does not have enough examples of fabricated sharp corners and therefore does not have the "intelligence" to fully correct them. Improved training layouts that include sharper features after fabrication will further improve the capabilities of the corrector model. Although perfectly reproducing the original sharp feature may not be feasible, the goal is still to get as close to it as possible.

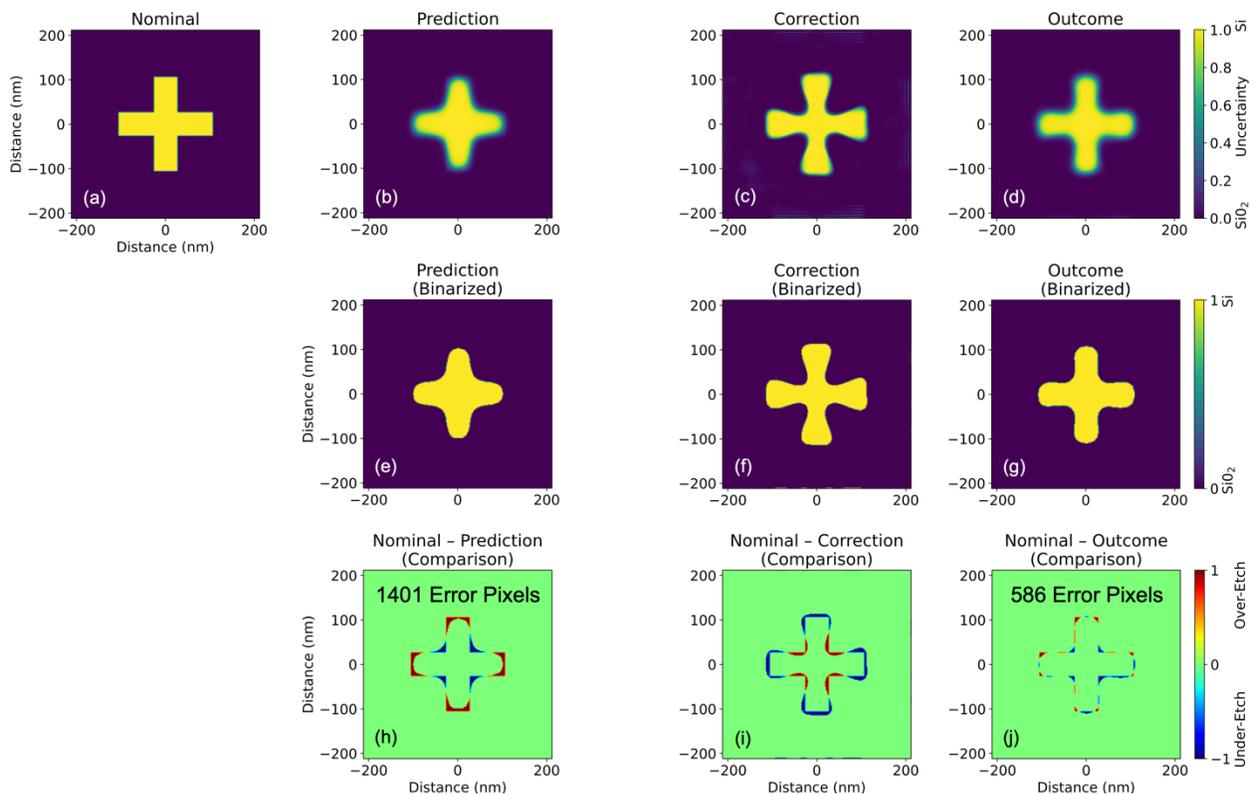

*Figure 5. Example of prediction and correction for a (a) simple silicon cross with 200×50 nm² crossings. The (b) prediction, (c) correction, and (d) prediction of correction (outcome), (e–g) their respective binarizations (binarization at 50% of their uncertainty regions), and (h–j) their respective comparisons with the nominal design, showing where there is loss or gain of silicon.*

## Results and Discussions

### Experimental Validation of Silicon Test Structure Correction

We have applied the correction model to a variety of structures and fabricated them to validate our methods. All structures in this section were corrected using the corrector model trained with a tandem architecture. We first present the fabrication results of two simple test structures for more clear illustration of the process variations. Figure 6 shows the test structures with and without correction. The first structure is a star shape of 200 nm across that experiences significant over-etching of its acute convex corners and light under-etching of its obtuse concave corners. This variation is severe for the non-corrected structure, where it looks closer to a pentagon than a star. When compared to the nominal star design, the non-corrected structure has 2,387 error pixels, while the corrected structure has only 1,063 error pixels—which is a reduction factor of 2.2. The second structure is a cross shape with 100×25 $nm^2$ crossings, where the 90° concave corners in the middle experience some under-etching, and the 90° convex corners experience severe over-etching. When compared to the nominal cross design, the non-corrected structure has 2,133 error pixels, while the corrected structure has only 825 error pixels—which is a reduction factor of 2.6. For a minimum accepted feature size of 60 nm specified by the nanofabrication facility, this structure represents significant miniaturization. Although these are only two examples and finding the exact degree of miniaturization requires further analysis and characterization, the results show a glimpse into future feature-limit-breaking designs for ultracompact, high-performance photonic circuits. The structures presented in this analysis are not optically useful, but their features are not uncommon to photonic device designs.

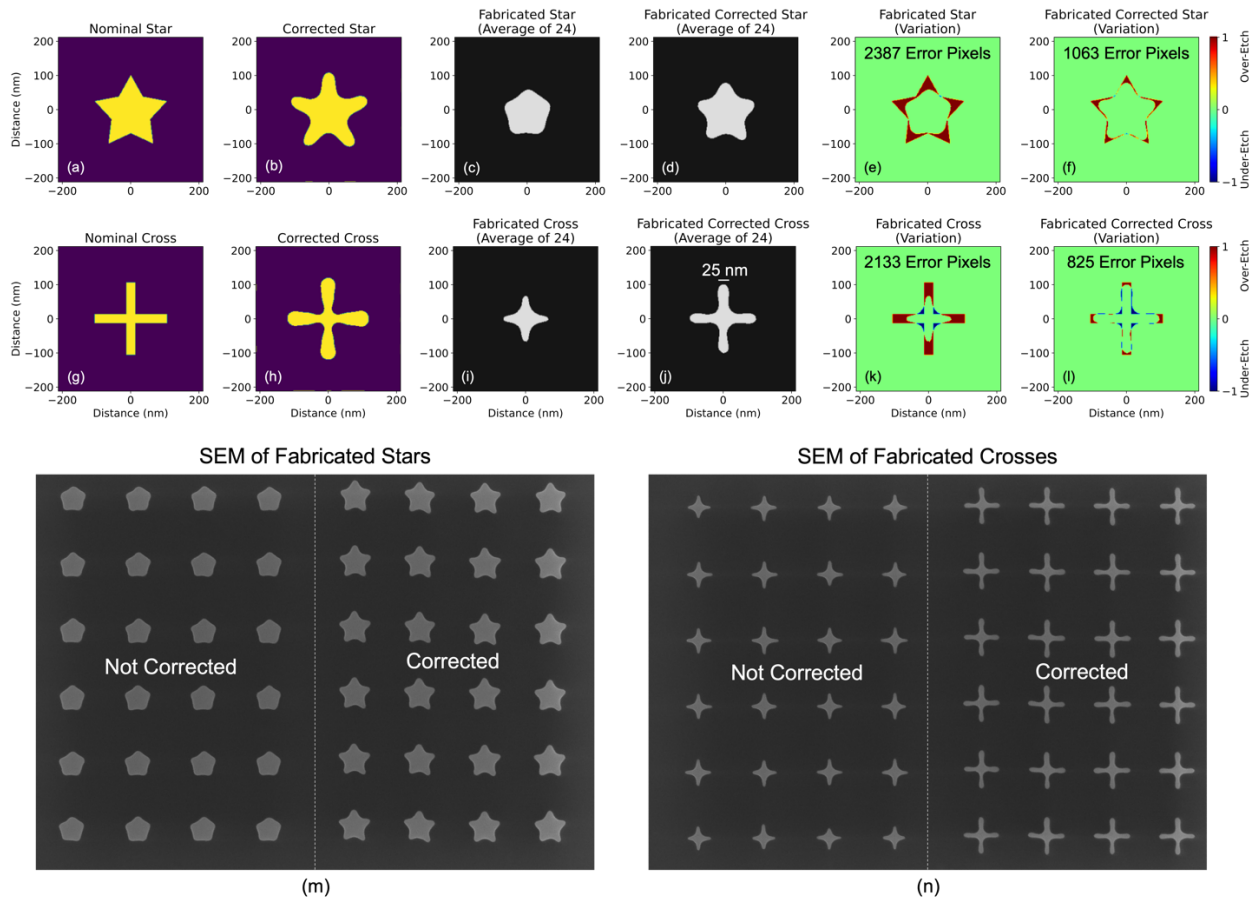

*Figure 6. Correction results of (a) a star structure and (g) a crossing structure. The structures' respective (b, h) corrections, (c, i) average shape of 24 fabrications, (d, j) average shape of 24 fabrications of their corrections, (e, k) comparisons of the average fabrication to the nominal, (f, l) comparisons of the average fabrication of the correction to the nominal, and (m, n) full SEM images of fabricated structures.*

## Photonic Device Correction

For a more practical demonstration of the (tandem-trained) corrector model, Figure 7 shows the prediction and correction results of a topologically optimized two-channel wavelength-division (de)multiplexer. This device is optimized with the LumOpt inverse design package (in 3D FDTD) from Ansys Lumerical[27] to maximize the demultiplexing of two wavelength ranges (1525 nm to 1535 nm and 1545 nm to 1555 nm) from one input waveguide to two output waveguides. A mesh size of 20 nm is used. To maximize the performance in a compact footprint of 5×3 µm$^2$, many small, complex features were generated, as shown in Figure 7a. The comparison in Figure 7g shows how the predicted deviations vary from the nominal design, including over/under-etching of sharp bends and washing away of small features. The prediction introduces 114,891 error pixels compared to the nominal. The comparison in Figure 7h shows where the correction adds or removes silicon to/from the nominal design. The comparison in Figure 7i shows how the prediction of the corrected structure (outcome) varies from the nominal design, which shows greater similarity than that without correction, having 73,650 error pixels—a reduction factor of 1.6.

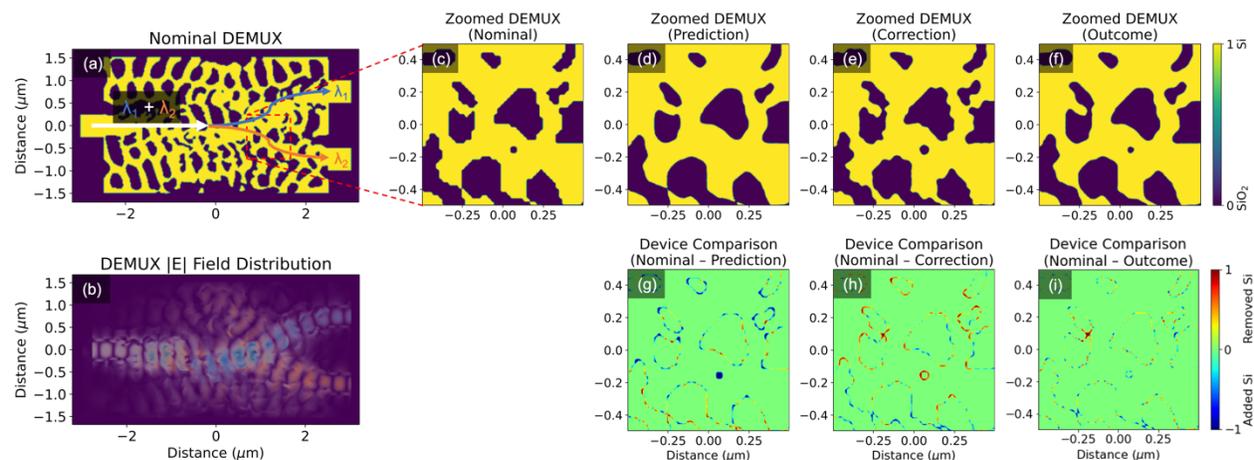

Figure 7. (a) The design of a topologically optimized wavelength-division (de)multiplexer and (b) the field profiles of its two operating wavelengths ($\lambda_1$ = 1530 nm and $\lambda_2$ = 1550 nm). Zoomed segments of (c) the nominal, (d) the prediction, (e) the correction, and (f) the predicted outcome. Zoomed comparisons between (g) the prediction and nominal, (h) the correction and nominal, and (i) the predicted outcome and nominal. For (g–i), green areas indicate no variation, red indicates removal of silicon, and blue indicates addition of silicon.

We then carried out optical simulations using the nominal design and the predicted structures, with and without correction. As shown by the 3D FDTD simulation results in Figure 8 and

Table 1, low insertion loss (IL) and crosstalk (XT) are achieved for both channels of the ideal, nominal design. A broadband transmission spectrum is also achieved for both channels of the nominal design, the prediction, and the prediction of the correction (outcome). As expected, the fabrication deviations of the prediction observed in Figure 7c led to a large shift in peak wavelength (Δλ) by more than 10 nm, increased IL by 2 dB, and increased XT by 10 dB. The corrected structure performs significantly better than the non-corrected design with Δλ of ony 1 nm. It should be noted that the IL and XT of the correction are marginally better than those of the nominal; however, this is inconsequential, as the nominal design was likely not fully optimized.

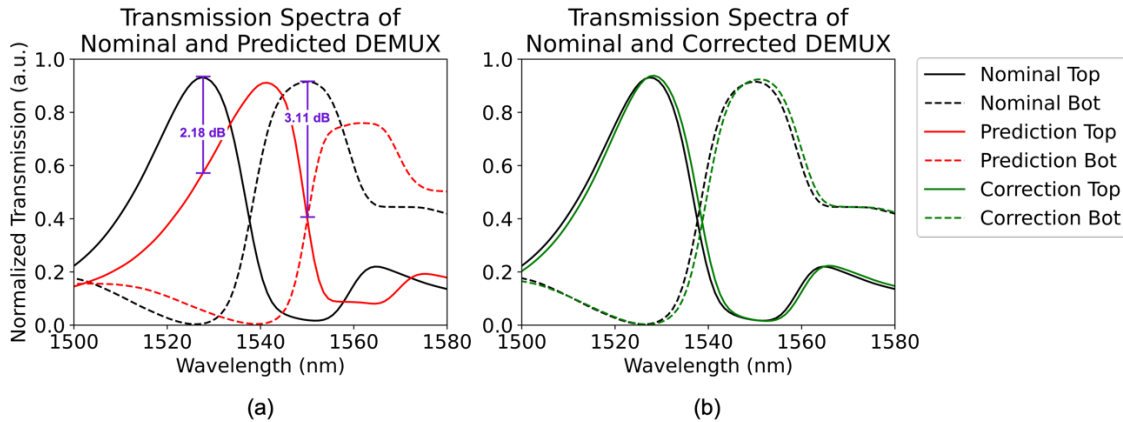

*Figure 8. 3D FDTD simulation results for the transmission spectra of (a) a fabrication-predicted topologically optimized wavelength-division (de)multiplexer, and (b) its corresponding prediction of correction. Simulation of the nominal design (black curves) is shown in both (a) and (b) for reference.*

*Table 1. Comparison of key simulation metrics for a topologically optimized two-channel wavelength-division (de)multiplexer, its prediction, and its prediction of correction.*

| Metric | Nominal | | Predicted Nominal | | Predicted Correction | |
|---|---|---|---|---|---|---|
| | Top | Bottom | Top | Bottom | Top | Bottom |
| **IL$_{peak}$ (dB)** | 0.31 | 0.38 | 2.49 | 3.49 | 0.29 | 0.34 |
| **XT$_{peak}$ (dB)** | -23.2 | -17.5 | -12.4 | -4.62 | -25.1 | -17.4 |
| **Δλ$_{peak}$ (nm)** | — | — | 14 | 11 | 1 | 1 |

The results in Figure 8 demonstrate how machine learning based predictor and corrector models work in a virtual fabrication environment for rapid, "fabless" prototyping. Complex designs can be verified prior to costly nanofabrication, and corrections can even be made automatically for fabrication-aware computer-aided design. Without modifying the design or nanofabrication processes, our corrector model significantly increases the performance of an existing design. We expect similar improvements can be made to all existing and future planar silicon photonic designs (to varying degrees of improvement defined by the complexity of the design). Furthermore, the device we corrected in this work has many features that violate the design rules of the foundry: their inevitable variations (along with smaller variations for features that do not violate design rules) cause performance degradation that our corrector model restores post-optimization. This can relax the strict optimization constraints that lead to less performant designs. With the expectation of using the corrector model in future designs, the design rules can be pushed beyond their limits for designs with even smaller features, ultimately enabling new, record-breaking levels of performance.

## Conclusion

The proliferation of silicon photonics is hindered by low tolerance to nanofabrication process variations. Next-generation design methods such as topology optimization have enabled impressive levels of performance and miniaturization at the expense of even lower fabrication tolerance. In this work, we demonstrated the use of deep learning to automatically correct for feature deviations in planar silicon photonic devices so that the fabricated outcome is closer to that of the nominal design. Only a modest set of SEM images is used to train the model for a commercial electron-beam lithography process, but the method can be readily adapted to any other similar process, such as deep UV lithography. Furthermore, the proposed corrector model adds further benefits by enabling smaller features than what are specified by the nanofabrication facility, opening the door to new, record-breaking designs without sacrificing reliability, adding significant computation, or changing the existing nanofabrication process.

# Funding
This work is supported by the National Research Council Canada Challenge Programs AI for Design (Grant AI4D-101-1) and High Throughput and Secure Network (Grant HTSN-219).